\documentclass[conference]{IEEEtran}
\IEEEoverridecommandlockouts
\usepackage{cite}
\usepackage{amsmath,amssymb,amsfonts}
\usepackage{algorithmic}
\usepackage{graphicx}
\usepackage{textcomp}
\usepackage{xcolor}
\usepackage{url}
\def\BibTeX{{\rm B\kern-.05em{\sc i\kern-.025em b}\kern-.08em
    T\kern-.1667em\lower.7ex\hbox{E}\kern-.125emX}}
\begin{document}

\title{Automatic Prompt Engineering \\with No Task Cues and No Tuning}

\author{
    \IEEEauthorblockN{Faisal Chowdhury\IEEEauthorrefmark{1}, Nandana Mihindukulasooriya\IEEEauthorrefmark{1}, Niharika S. D'Souza\IEEEauthorrefmark{1}, Horst Samulowitz\IEEEauthorrefmark{1}, \\Neeru Gupta\IEEEauthorrefmark{2}, Tomasz Hanusiak\IEEEauthorrefmark{2}, Michal Kapitonow\IEEEauthorrefmark{2}}
    \IEEEauthorblockA{\IEEEauthorrefmark{1}IBM Research
    \\mchowdh@us.ibm.com, nandana@ibm.com, Niharika.DSouza@ibm.com, samulowitz@us.ibm.com}
    \IEEEauthorblockA{\IEEEauthorrefmark{2}IBM
    \\guptaneeru@us.ibm.com, T.Hanusiak@pl.ibm.com, michal.kapitonow@ibm.com}
}
 
\maketitle

\begin{abstract}
This paper presents a system for automatic prompt engineering that is much simpler in both design and application and yet as effective as the existing approaches. It requires no tuning and no explicit clues about the task. 
We evaluated our approach on cryptic column name expansion (CNE) in database tables, a task which is critical for tabular data search, access, and understanding and yet there has been very little existing work. We evaluated on datasets in two languages, English and German. This is the first work to report on the application of automatic prompt engineering for the CNE task. To the best of our knowledge, this is also the first work on the application of automatic prompt engineering for a language other than English. 
\end{abstract}


\section{Introduction}
The advent of LLMs, especially after the impressive demonstrations of ChatGPT, led to a naive belief in many customers/companies that a new dawn of business intelligence is on the horizon which would require very limited effort to exploit the LLMs on their usecases. It turned out that, more often than not, the downstream performance of LLMs is heavily coupled with the quality of the prompt used to instruct the model. Studies have shown that large language models (LLMs) are quite sensitive to minor variations in prompt phrasing. For example, the addition, removal, or reordering of just a few tokens can lead to significant differences in task performance~\cite{b11}. Generic prompts do not typically produce good responses and the most effective prompts are almost always handcrafted by humans. This makes prompt curation a labour-intensive iterative process involving a substantial amount of manual experimentation. This process of optimizing the prompt language to elicit the best possible performance is referred to as ``prompt engineering". Given the human effort involved in the practice, prompt engineering techniques are often brittle, non-transferable, and suffer from scalability issues ~\cite{b4}.

Currently, prompt engineering is more of an art than a science, as a delicate balance is required in the design process to ensure clarity and specificity in the prompts, avoid ambiguity, and steer appropriate behavior to obtain the desired output. More importantly, the human labor involved needs to be repeated whenever the underlying LLM is changed, since previously optimized prompts may no longer yield optimal results. For example, this applies when we switch to a different model, to a different parameter-size variant of the same model, or upgrade to new versions of the LLM trained using different strategies/additional data. This is further complicated when the prompt has to take into account domain specificity or a different language (direct translations of prompts often produce poor results~\cite{b12}). 

Automatic prompt engineering and optimization are  recent trends. The core idea is that an LLM, not a human, is tasked with generating task-specific prompts where the task is presented via output demonstrations (with some given examples). The corresponding LLM generates several instruction candidates, either via direct inference or a recursive process driven by scoring metrics. The LLM executes these instructions, and the best instruction improving the scoring metric is retained. Automatic prompt engineering is envisaged to be applied to any task that is solved by prompting LLMs.

However, existing approaches often require an initial human written task specific prompt. They also need extensive number of expensive LLM calls during the optimization process. In addition, they are not easily adoptable due to implementation complexities. Also, all of them require evaluation data for their scoring metrics to work to pick the best possible prompt.

In this paper, we propose a system that is based on a new simple yet effective approach for automatic prompt engineering that uses only a few examples and controlled randomized sampling to generate the best possible prompt.

To be more precise, the advantages of our approach with respect to known solutions are as follows:
\begin{itemize}
    \item Simple approach that can be easily adopted (even for non-English languages) and works with only a few given examples. 
    \item No need of separate training, validation and test data.
    \item No need of initial seed prompt or task specific cues.
    \item No additional LLM calls for scoring or ranking.
\end{itemize}

 We evaluated our approach on Cryptic column name expansion (CNE) in database tables. We used three evaluation datasets. Two in English and one in German. 

To the best of our knowledge, this is the first work on the application of automatic prompt engineering for a language other than English. 

Furthermore, as part of this work, we are going to make available a dataset in German for the CNE task.
 
We demonstrate that our system is as good as or better than the other existing system but much easier to adopt and much simpler in design.

\section{Task: Cryptic Column Name Expansion (CNE) in Tables}
Tables in real world customer database often have Short or cryptic column names such as following:

\begin{quote}
    (English example)\\
    \texttt{\small \{"table": "Customer Account", "columns": ["DISCOUNT\_PCT\_APPLIC", "CURRENT\_BAL\_AMT"]\}}\\
    (German example)\\
    \texttt{\small \{"table": "vbuk", "columns": ["mandt", "vbeln", "gbstk", "vbtyp", "aedat"]\}}\\
\end{quote}

The goal of the CNE task is to expand such column names and produce an output like following:

\begin{quote}
(English example)\\
\texttt{\small \{"table": "Customer Account", "columns": \{"DISCOUNT\_PCT\_APPLIC": "Discount Percentage Applicable", "CURRENT\_BAL\_AMT": "Current Balance Amount"\}\}}\\
    (German example)\\
    \texttt{\small \{"table": "vbuk", "columns": \{"mandt": "Mandant", "vbeln": "Vertriebsbelegnummer eines CAS-Kontaktes", "gbstk": "Gesamtbearbeitungsstatus des Vertriebsbeleges", "vbtyp": "Vertriebsbelegtyp", "aedat": "Datum der letzten Änderung"\}\}}
\end{quote}

Column names are often not abbreviated in isolation but in the context of the table and other column names in the corresponding table.

\cite{b21} showed that abbreviated column names makes it challenging for end users to search and retrieve relevant data for many table-related tasks. One of the widely used human-labeled text2SQL benchmark, the Spider dataset \cite{b22}, contains 6.6\% of abbreviated column names. \cite{b21} observed over ten percentage points in performance degradation on the Spider dataset due to simple changes in abbreviated column names.  Abbreviated column names also effect table question answering (QA) \cite{b23}.

To the best of our knowledge, \cite{b20} is the only existing peer-reviewed work so far for this task that provided experimental results. As they noted, expanding column names has other beneficial aspects such as increased readability of tables (especially when complex or technical data is present), disambiguating between tables with similar column names but different meanings, improved the efficacy of keyword based searches for discovering related tables, etc. As a solution for this task, they semi-automatically generated a large training dataset in English and then fined-tuned an LLM. In contrast, our proposed system requires no training. 

\section{Related Work on Automatic Prompt Optimization and Engineering}

Prompt optimization aims to refine and tune of an existing or original prompt to improve performance across multiple runs or datasets. Whereas the goal of prompt engineering is to design a prompt structure from scratch, often by using techniques like few-shot prompting.

\cite{b2} recently provided a through evaluation of of various automatic prompt engineering and optimization approaches for the task of triple extraction from text.

The works of~\cite{b5} provide a comprehensive taxonomy of automatic prompt optimization frameworks which refine prompts with no or minimal human intervention. Broadly, these methods can be categorized along multiple dimensions. These include:  optimization space (i.e. discrete text-based vs. soft prompting or gradient-based), optimization targets (i.e. instructions vs. examples),  optimization objective (i.e. task performance, safety, or generalizability), operators used to generate new prompts (e.g. purely model-based vs. iterative refinement of example prompts), and iterative search strategies (e.g. Evolutionary vs. Monte Carlo search). Below we mention a few notable of these approaches.

A popular prompt optimization system is the work of~\cite{b15} called DSPy that frames prompt as a declarative, compiler-driven optimization task. Gradient-based approaches such  as TextGrad~\cite{b6} use gradient descent-like algorithms to optimize prompt embeddings according to a predefined performance objective~\cite{b17}.  

With regard to automatic prompt engineering, \cite{b3} proposed one of the earliest work via in-context learning that they termed as ``instruction induction''.  To elicit models to generate instructions, they created a meta-prompt presenting instruction induction as a challenge puzzle. Their meta prompt was --\\
\hrule
\begin{quote}
\small
\texttt{I gave a friend an instruction and five inputs. The friend read the instruction and wrote an output for every one of the inputs. Here are the input-output pairs:}

\texttt{Input: ...} \\
\texttt{Output: ...}\\
\texttt{Input: ... }\\
\texttt{Output: ...}\\
\texttt{...}\\

\texttt{The instruction was <COMPLETE>}
\end{quote}
\hrule

~\\

This approach uses the greedy decoding algorithm to generate a single prompt, effectively avoiding the need to design a mechanism for selecting the best possible prompt of non-greedy decoding was used during inference.

Automated Prompt Engineer (APE)~\cite{b10} extends the approach of \cite{b3} with a search and selection process through a pool of instruction candidates proposed by an LLM to maximize a chosen score function. 

A complementary approach that uses reinforcement learning-based strategies is OIRL~\cite{b13}, which model the interaction between the query-prompt pair via a reward model for proposing and evaluating candidate prompts suited for arithmetic reasoning tasks. In contrast, meta-Prompting~\cite{b14} uses structural and syntactical aspects of tasks to create general prompts that guide the generation of task-specific prompts. 

\section{Proposed System}

Similar to \cite{b3}, our approach uses a generic meta-prompt (independent of task) with a few example input-output pairs. However, unlike \cite{b3}, we do not use greedy decoding. We use multinomial sampling decoding. Also, unlike ~\cite{b10}, our scoring function does need additional LLM calls to rank the generated candidate prompts. 

There are two steps in our proposed approach.

\subsection{Step 1: Generate candidate prompts specific for the target task}

Given a few (8-10) example pairs of input and output for the target task, the system creates 3 samples:

\begin{enumerate}
    \item Randomly choose a small (4-5) examples from the list. Let’s call it sample A
    \item Randomly choose another small (4-5) examples from the list that are not in sample A. Let’s call it sample B.
    \item Randomly choose half of the examples from sample A and half from sample B. \item Put them in a new list. Let’s call it sample C.
\end{enumerate}

Each of the samples above is combined with the meta-prompt to construct three input prompts. Our \textbf{task agnostic meta-prompt} for English is --\\
\hrule
\begin{quote}
\small
\texttt{I gave a friend an instruction. Based on the instruction he produced the following input and output pairs:}

\texttt{Input: ...} \\
\texttt{Output: ...}\\
\texttt{Input: ... }\\
\texttt{Output: ...}\\
\texttt{...}\\

\texttt{Complete the following text.\\The instruction was to <COMPLETE>}
\end{quote}
\hrule

~\\~\\Our \textbf{task agnostic meta-prompt} for German is --\\
\hrule
\begin{quote}
\small
\texttt{Ich gab einem Freund eine Anweisung. Danach erzeugte er die folgenden Eingabe und Ausgabepaare:}

\texttt{Input: ...} \\
\texttt{Output: ...}\\
\texttt{Input: ... }\\
\texttt{Output: ...}\\
\texttt{...}\\

\texttt{Vervollständigen Sie den folgenden Satz. Die Anweisung lautete: <COMPLETE>}
\end{quote}
\hrule
~\\

For each of the three input prompts, the system generates N (~10) number of candidate prompts using an LLM with multinomial sampling decoding.

\subsection{Step 2: Rank generated candidate prompts}
For each generated prompts, the system calculates similarity scores with respect to every other generated prompts, sums the scores and then averages it. This would be the score for this particular generated instruction. We use an approximate string similarity algorithm called Jaro-Winkler similarity\footnote{https://en.wikipedia.org/wiki/Jaro–Winkler\_distance}.

The system then ranks all the generated prompts in descending order according to their scores. The system outputs the top ranked prompt as the desired prompt for the target task.

Our choice of Jaro-Winkler similarity instead of LLM based scoring is primarily due to avoid additional LLM calls both for a faster output generation time and also to save cost of customers for making the LLM calls and the tokens generated due to those calls.

In addition, our manual analysis showed that an approximate string similarity approach like Jaro-Winkler similarity is able to filter out candidate prompts which are either too verbose or contain inconsistent task instructions.   

\section{Evaluation Results}
We evaluate our approach against the following existing approaches: Instruction Induction (shortened as InstInduc)~\cite{b3}, APE Zeroshot~\cite{b10}, TextGrad~\cite{b6}, and DSPy~\cite{b15}.

For consistency, we used the 
Llama-3.3-70B-Instruct\footnote{https://huggingface.co/meta-llama/Llama-3.3-70B-Instruct} model in all the systems (including ours) to generate prompts.

\begin{figure}[htbp]
  \centering
  \includegraphics[width=\linewidth]{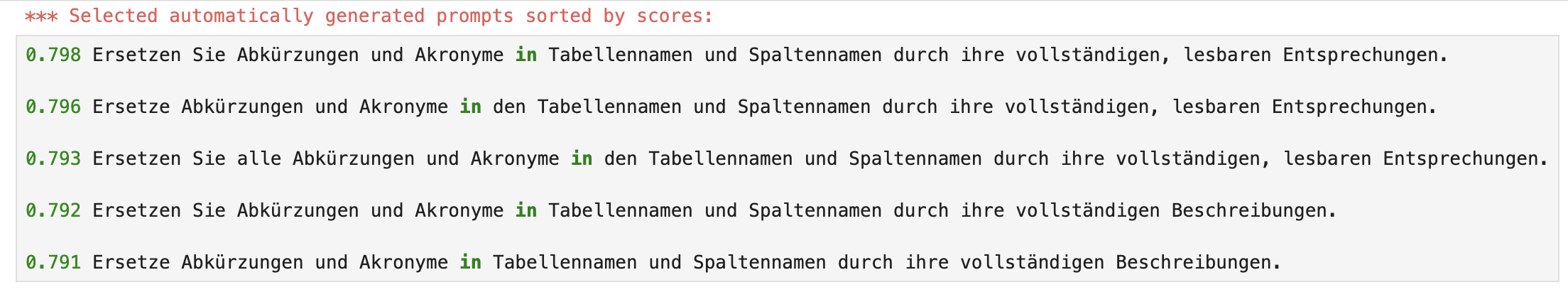}
  \caption{Output of our system for the German CNE task given only few examples as input.}
  \label{fig:ToolImpactAndLoop}
\end{figure}

~\\\textbf{CNE English datasets:} We use 3 datasets for English: \textsc{CSTINSIGHT}~\cite{b1} which is a database of customer insights, \textsc{CDO\_435} which is a database from the chief data officer of one of our customers, and \textsc{Tele\_1186} which is a database from a telecommunication customer.

The \textsc{CSTINSIGHT} contains 25 tables and 519 cryptic columns. Since both DSPy and TextGrad need separate training and validation data, we used 19 of these tables for training and 6 of them for validation for those 2 systems.

For our system as well as for InstInduc, APE and APE Zeroshot, we used only 10 of these \textsc{CSTINSIGHT} tables as examples (since these 3 systems do not need any tuning).

The \textsc{CDO\_435} and \textsc{Tele\_1186} datasets were used for testing. \textsc{CDO\_435} has 11 tables and 424 cryptic columns, whereas \textsc{Tele\_1186} has 46 tables and 1140 cryptic columns.

~\\\textbf{CNE German dataset:} We created this dataset from the publicly available \textsc{SAP BigQuery Dataset}\footnote{https://www.kaggle.com/datasets/mustafakeser4/sap-dataset-bigquery-dataset}. Using publicly available \textsc{SAP ERP 6.0}\footnote{https://sap.erpref.com/?schema=ERP6EHP7} and \textsc{SAP NetWeaver 7.4}\footnote{https://help.sap.com/doc/saphelp\_nw74/7.4.16/de-DE/52/367e53f33d6359e10000000a174cb4/frameset.htm} documentation, we selected only those tables in this database that contains at least 5 cryptic column names. 

We are going to release this dataset under the same MIT license of the \textsc{SAP BigQuery Dataset}. This dataset contains total 23 tables and 283 cryptic columns.

We used 15 of these 23 tables as test data and remaining 8 tables as few-shot examples for prompt generation for our system, InstInducAPE and APE Zeroshot. For DSPy and TextGrad, 3 of these 8 examples are used for training and 5 used for validation.

\begin{table}[htbp]
\caption{Results on the German and English CNE datasets. Accuracy was calculated using Jaro-Winkler similarity ($>=$0.85).}
\begin{center}
\begin{tabular}{c | c | c | c}
    \hline
    \textbf{System} & \textbf{\textsc{German SAP}} & \textbf{\textsc{CDO\_435}} & \textbf{\textsc{Tele\_1186}} \\
    \textbf{name} & \textbf{(Accuracy)} & \textbf{(Accuracy)} & \textbf{(Accuracy)} \\
    \hline
    InstInduc & 21.08 & 48.11 & 46.77 \\
    APE Zeroshot & 41.13 & 79.95 & 68.92 \\                         
    TextGrad & 48.11 & 72.17  &  59.04  \\
    DSPy & \bf 51.89 & 69.34 &  \bf 75.00                       \\
    Our system & \bf 51.89 & \bf 82.61 &  70.73     \\
    \hline
  \end{tabular}
\label{tbl:cne}
\end{center}
\end{table}

As we see in Table \ref{tbl:cne}, with respect to the more complex DSPy system (which requires tuning as well initial task specific cues), our system achieves similar results for the German SAP dataset, performs better on the English \textsc{CDO\_435} and obtains lower results on the English \textsc{Tele\_1186} dataset. However, our system outperforms all the other existing approaches we tested. Particularly, among the 185 cryptic columns in the 15 test tables of the German SAP dataset, both our system and DSPy got 89 of them wrong. Interestingly, for 37 of these wrong predictions, our system and DSPy both predicted the same expansion. 

\section{Conclusion}

We presented a system for automatic prompt engineering that is simple, language adaptable, task agnostic and yet can produce as good or better results than more complex existing systems on the important but less explored CNE task. We obtained similar results for the triple extraction task \cite{b2} but could not include due to space limitation.


\begin{thebibliography}{00}
\bibitem{b1} J. Baek, H. Samulowitz, O. Hassanzadeh, D. Subramanian, S. Shirai, A. Gliozzo and D. Bhattacharjya, ``Knowledge Base Construction for Knowledge-Augmented Text-to-SQL'', Findings of the 63rd Annual Meeting of the Association for Computational Linguistics, July 2025.
\bibitem{b2} N. Mihindukulasooriya, N. S. D’Souza, F. Chowdhury and H. Samulowitz, ``Automatic Prompt Optimization for Knowledge Graph Construction: Insights from an Empirical Study'', Proceedings of the VLDB 2025 Workshop: LLM+Graph, September 2025.
\bibitem{b3} O. Honovich, U. Shaham, S. R. Bowman and O. Levy, ``Instruction Induction: From Few Examples to Natural Language Task Descriptions'', Proceedings of the 61st Annual Meeting of the Association for Computational Linguistics (ACL), pp. 1935--1952, July 2023.
\bibitem{b4} S. Arora, A. Narayan, M. F. Chen, L. Orr, N. Guha, K. Bhatia, I. Chami, C. Re, ``Ask Me Anything: A simple strategy for prompting language models'', Proceedings of the 11th International Conference on Learning Representations (ICLR), May 2023.
\bibitem{b5} W. Cui, J. Zhang, Z. Li, H. Sun, D. Lopez, K. Das, B. Malin and S. Kumar, ``Automatic Prompt Optimization via Heuristic Search: A Survey'', Findings of the 63rd Annual Meeting of the Association for Computational Linguistics (ACL), pp. 22093-–22111
July, 2025.
\bibitem{b6} M. Yuksekgonul, F. Bianchi, J. Boen, S. Liu, P. Lu, Z. Huang, C. Guestrin and J. Zou, James, ``Optimizing generative AI by backpropagating language model feedback'', Nature (volume 639), pp. 609--616, March 2025.
\bibitem{b9} H. Hugo, T. Lavril, G. Izacard, X. Martinet, M. Lachaux, and T. Lacroix, ``Llama: Open and efficient foundation language models'', arXiv preprint arXiv:2302.13971, February 2023.
\bibitem{b10} Y. Zhou, A. I. Muresanu, Z. Han, K. Paster, S. Pitis, H. Chan, and J. Ba, ``Large language models are human-level prompt engineers'', Proceedings of the 11th International Conference on Learning Representations (ICLR), May 2023.
\bibitem{b11} P. Liu, W. Yuan, J. Fu, Z. Jiang, H. Hayashi, G. Neubig, ``Pre-train, Prompt, and Predict: A Systematic Survey of Prompting Methods in Natural Language Processing'', ACM Computing Surveys, Volume 55, Issue 9, pp. 1--35, January 2023.
\bibitem{b12} I. Mondshine, T. Paz-Argaman and R. Tsarfaty,Beyond {E}nglish: The Impact of Prompt Translation Strategies across Languages and Tasks in Multilingual {LLM}s'', Proceedings of the 8th Workshop on Technologies for Machine Translation of Low-Resource Languages, May 2025.
\bibitem{b13} H. Sun, A. H{\"u}y{\"u}k and M. van der Schaar, ``Query-Dependent Prompt Evaluation and Optimization with Offline Inverse RL'', Proceedings of the 12th International Conference on Learning Representations, 2024.
\bibitem{b14} M. Suzgun and A. T. Kalai, ``Meta-Prompting: Enhancing Language Models with Task-Agnostic Scaffolding'', arXiv preprint arXiv:2401.12954, January, 2024.
\bibitem{b15} O Khattab, A. Singhvi, P. Maheshwari, Z. Zhang, K. Santhanam, S. Vardhamanan, S. Haq, A. Sharma, T. Joshi, H. Moazam, H. Miller, M. A.Zaharia, C. Potts, ``DSPy: Compiling Declarative Language Model Calls into Self-Improving Pipelines'', Proceedings of the Workshop on robustness of zero/few-shot learning in foundation models, 2023
\bibitem{b17} Y. Wen, N. Jain, J. Kirchenbauer, M. Goldblum, J. Geiping, T. Goldstein, ``Hard Prompts Made Easy: Gradient-Based Discrete Optimization for Prompt Tuning and Discovery'', Proceedings of the 37th Conference on Neural Information Processing Systems (NeurIPS), 2023.
\bibitem{b19} M. Josifoski, M. Sakota, M. Peyrard and R. West, ``Exploiting Asymmetry for Synthetic Training Data Generation: {S}ynth{IE} and the Case of Information Extraction'', Proceedings of the 2023 Conference on Empirical Methods in Natural Language Processing, December, 2023.
\bibitem{b20}  J. Zhang, Z. Shen, B. Srinivasan, S. Wang, H. Rangwala and G. Karypis, ``NameGuess: Column Name Expansion for Tabular Data'',  Proceedings of the 2023 Conference on Empirical Methods in Natural Language Processing, pp 13276–-13290, December, 2023.
\bibitem{b21} T. Xie, C. H. Wu, P. Shi, R. Zhong, T. Scholak, M. Yasunaga, C. S. Wu, M. Zhong, P. Yin, S. I. Wang, V. Zhong, B. Wang, C. Li, C. Boyle, A. Ni, Z. Yao, D. Radev, C. Xiong, L. Kong, R. Zhang, N. A. Smith, L. Zettlemoyer, and T. Yu, ``Unifiedskg: Unifying and multi-tasking structured knowledge grounding with text-to-text language models'', Proceedings of the 2022 Conference on Empirical Methods in Natural Language Processing (EMNLP), pp.. 602–631, December, 2022. 
\bibitem{b22} T. Yu, R. Zhang, K. Yang, M. Yasunaga, D. Wang, Z. Li, J. Ma, I. Li, Q. Yao, S. Roman, Z. Zhang and D. R. Radev, ``Spider: A large-scale human-labeled dataset for complex and cross-domain semantic parsing and text-to-sql task'', Proceedings of the 2018 Conference on Empirical Methods in Natural Language Processing (EMNLP), October, 2018.
\bibitem{b23} P. Yin, G. Neubig, W. Yih and S. Riedel, ``TaBERT: Pretraining for joint understanding of textual and tabular data'', Proceedings of the 58th Annual Meeting of the Association for Computational Linguistics (ACL), pp. 8413–-8426, July, 2020.

\end{thebibliography}
\end{document}